# Speeding Up Planning in Markov Decision Processes via Automatically Constructed Abstractions

**Alejandro Isaza** and **Csaba Szepesvári** and **Vadim Bulitko** and **Russell Greiner**
Department of Computing Science, University of Alberta
Edmonton, Alberta, T6G 2E8, CANADA
{isaza,szepesva,bulitko,greiner}@cs.ualberta.ca

## Abstract

In this paper, we consider planning in stochastic shortest path (SSP) problems, a subclass of Markov Decision Problems (MDP). We focus on medium-size problems whose state space can be fully enumerated. This problem has numerous important applications, such as navigation and planning under uncertainty. We propose a new approach for constructing a multi-level hierarchy of progressively simpler abstractions of the original problem. Once computed, the hierarchy can be used to speed up planning by first finding a policy for the most abstract level and then recursively refining it into a solution to the original problem. This approach is fully automated and delivers a speed-up of two orders of magnitude over a state-of-the-art MDP solver on sample problems while returning near-optimal solutions. We also prove theoretical bounds on the loss of solution optimality resulting from the use of abstractions.

## 1 Introduction and Motivation

We focus on planning in stochastic shortest path problems (the problem of reaching some goal state under uncertainty) when planning time is critical — a situation that arises, for instance, in path planning for agents in commercial video games, where map congestions are modeled as uncertainty of transitions. Another example is path planning for multi-link robotic manipulators, where the uncertainty comes from unmodeled dynamics as well as sensor and actuator noise. More specifically, we consider the problem of finding optimal policies in a sequence of stochastic shortest-path problems (Bertsekas & Tsitsiklis, 1996), where the problems share the same dynamics and transition costs, and differ only in the location of the goal-state.

When the state space underlying the problems is sufficiently large, exact planning methods are unable to deliver a solution within the required time, forcing the user to resort to approximate methods in order to scale to large domains. Exploiting the fact that multiple planning problems share the same dynamics and transition costs, we build an abstracted representation of the shared structure where planning is faster, then map the individual planning problem into the abstract space and derive a solution there. The solution is then refined back into the original space.

In a related problem of path planning under real-time constraints in *deterministic* environments (e.g., Sturtevant, 2007), a particularly successful approach is implemented in the PR LRTS algorithm (Bulitko, Sturtevant, Lu, & Yau, 2007), which builds an abstract state space by partitioning the set of states into cliques (i.e., each state within each cluster is connected to each other state in that cluster with a single action). Each such cluster becomes a single abstract state. Two abstract states are connected by an abstract transition if there is a pair of non-abstract states (one from each abstract state) connected by a single action. The resulting abstract space is smaller and simpler, yet captures some of the structure of the original search problem. Thus, an abstract solution can be used to guide and constrain the search in the original problem, yielding a significant speed-up. Further speed-ups can be obtained by building abstractions on top of abstractions, which creates a hierarchy of progressively smaller abstract search spaces. PR LRTS can then be tuned to meet strict real-time constraints while minimizing solution suboptimality.

Note that state cliques produced by PR LRTS make good abstract states because landing anywhere in such a cluster puts the agent a single action away from any other state in the clique. This also means that the costs of the resulting actions are similar, and that the cost of a single action is negligible compared with the cost of a typical path. Finally, any (optimal) path in the original problem can be closely approximated at the abstract level, as an agent following an (optimal) path has to traverse from cluster to cluster. Since all neighboring clusters are connected in the abstract problem, it is always possible to find a path in the abstract problem that is "close" to the original path.

Given the attractive properties or PR LRTS, it is natural to ask whether the ideas underlying it can be extended to *stochastic* shortest path problems, with *arbitrary* cost structures.

In a stochastic problem, the result of planning is a closed loop policy that assigns actions to states. A successful ab-straction must be suitable for approximating the execution trace of an optimal policy. Imagine that clustering has been done in some way. The idea is again to have abstract actions that connect neighboring clusters cheaply — that is, the system should not produce expensive connections.

Intuitively, we want to connect one cluster to another if, from any state of the first cluster, we can reliably get to some state of the second cluster at roughly a fixed cost (the same for any state in the first cluster). This way, "simulating" a policy of the original problem becomes possible at a small additional cost (the meaning of simulation will become clear later). This means that a connection between clusters is implemented by a policy with a specific set of initial states that brings the agent from any state of the source cluster to some state of the target cluster. We will use options (Sutton, Precup, & Singh, 1999) for such policies, and choose clusters to allow such policies for any two neighboring clusters. Thus, it is natural to look for clusters of states that allow one to reliably simulate any trajectory from any of the states to any other state.

Finally, we need an extra mechanism, the "goal approach", that deals with the challenge of reaching the base-level goal itself from states that are close to the goal. Thus, our planner first plans in the abstract space to reach the "goal cluster". After arriving at some state of the "goal approach region", the planner then uses the "goal-approach policy" that, with high probability, moves the agent to the goal state itself. These ideas form the core of our algorithm.

The three major contributions of the paper are: (i) a novel theoretical analysis of option-based abstractions, (ii) an effective algorithm for constructing high-quality option-based abstractions, and (iii) experimental results demonstrating that our algorithm performs effectively over a range of problems of varying size and difficulty.

Section 2 formally describes our problem, and provides the theoretical underpinning of our approach. Section 3 then presents our algorithm for automatically building options-based abstractions, and Section 4, our planning algorithm that uses these abstractions. Section 5 empirically evaluates this approach, in terms of both efficiency and effectiveness (suboptimality). Finally, Section 6 summarizes related work.

## 2 Problem Formulation and Theory

This section formally defines stochastic shortest path problems and the abstractions that we will consider. It also presents a theoretical result that characterizes the relationship between the performance of abstract policies and policies of the original problem.

**Definition 1** A *Markov Decision Process* (MDP) is defined by a finite state space $X = \{1, \ldots, n\}$; a finite set of actions $A(x)$ for each state $x \in X$; transition probabilities $p(y|x, a) \in [0, 1]$ that correspond to the probability that the next state is $y$ when action $a$ is applied in state $x$; immediate cost $c(x, a, y) \in \Re$ for all $x, y \in X$ and all $a \in A(x)$ and a discount factor $\gamma \in (0, 1]$. The MDP is undiscounted if $\gamma = 1$. An action $a \in \cup_{x \in X} A(x)$ is called admissible in state $x$ if $a \in A(x)$.

**Definition 2** A *(generic) policy* is a mapping that assigns to each history $(x_0, a_0, c_0, \ldots, x_{t-1}, a_{t-1}, c_{t-1}, x_t)$ an action admissible in the most recent state $x_t$. In general, a mapping that maps possible histories to some set is called a *history dependent mapping*.

Under mild conditions, it suffices to consider only stationary, deterministic policies (Bertsekas & Tsitsiklis, 1996), on which we will focus:

**Definition 3** A stationary and deterministic *policy* $\pi$ is a mapping of states to actions such that $\pi(x) \in A(x)$ holds for any state $x \in X$.

In what follows, we will use "policy" to mean stationary and deterministic policies, unless otherwise mentioned.

The expected cost of policy $\pi$ when the system starts in state $x_0$ is $v_\pi(x_0) = \mathbb{E}\left[\sum_{t=0}^{\infty} \gamma^t c(X_t, \pi(X_t), X_{t+1})\right]$ where $X_t$ is a Markov chain with $\mathbb{P}\left(X_{t+1} = y | X_t = x\right) = p(y|x, \pi(x))$. The function $v_\pi$ is called the value-function underlying policy $\pi$.

One "solves" an MDP by finding a policy that minimizes the cost from every state, simultaneously. In this paper we deal only with stochastic shortest path problems, a subclass of MDPs. In these MDPs the problem is to get to a goal state with the least cost:

**Definition 4** A finite *stochastic shortest path* (SSP) problem is a finite undiscounted MDP that has a special state, called the goal state $g$, such that $\forall a \in A(g)$, we have $p(g|g, a) = 1$ and $c(g, a, g) = 0$ and the immediate costs for all the other transitions are positive.

Consider a finite SSP $(X, A, p, c)$. Let $\pi$ be a stationary policy. We say that this policy is *proper* if it reaches the goal state $g$ with probability one, regardless of the initial state. Let $T_\pi : \mathbb{R}^X \to \mathbb{R}^X$ be the policy's evaluation operator:

$$(T_\pi v)(x) = \sum_{y \in X} p(y|x, \pi(x)) \left[c(x, \pi(x), y) + v(y)\right].$$

Bertsekas and Tsitsiklis (1996) prove that $T_\pi$ is a contraction with respect to a weighted maximum norm, $\|\cdot\|_{w,\infty}$, with some positive weights, $w \in \mathbb{R}_+^X$, where $\|v\|_{w,\infty} = \max_x |v(x)|/w(x)$. In particular, $w(x)$ can be chosen to be the expected number of steps until $\pi$ reaches the goal state when started from $x$. The contraction coefficient of $T_\pi$, $\gamma_\pi$, satisfies $1/(1-\gamma_\pi) = \max_{x \in X} w(x)$. Thus, $1/(1-\gamma_\pi)$ is the maximum of the expected number of steps to reach the goal state, or in other words, the maximum expected time policy $\pi$ spends in the MDP (cf. Prop 2.2 in Bertsekas & Tsitsiklis, 1996).

We adopt the notion of options from Sutton et al. (1999):

**Definition 5** An *option* is a triple $(\pi, I, \psi)$, where $I \subset X$ is the set of initial states, $\pi$ is a (generic) policy that is defined for histories that start with a state in $I$ and $\psi$ is a history dependent mapping with range $\{0,1\}$, called the terminating condition. We say that the terminating condition fires when $\psi(h_t) = 1$. Let $T > 0$ denote the random time when the terminating condition fires for the first time while following $\pi$. (Note that $T = 0$ is not allowed.) We assume that $\mathbb{P}(T < +\infty) = 1$, independent of the initial state when the policy $\pi$ is started (i.e., the option terminates in finite time with probability one).

As suggested in the introduction, an abstraction is a way to group states and the abstract actions correspond to options:

**Definition 6** We say that the MDP $(\tilde{\mathcal{X}}, \tilde{\mathcal{A}}, \tilde{p}, \tilde{c})$ is an *option-based abstraction* of $(X, A, p, c)$, if there exists a mapping, $S : \tilde{\mathcal{X}} \to 2^X$ specifying the states $S(\tilde{x}) \subset X$ that correspond to an abstract state $\tilde{x} \in \tilde{\mathcal{X}}$, a set $\Pi$ of options abstracting the actions of the MDP and a mapping $\Psi : \cup_{\tilde{x}\in\tilde{\mathcal{X}}} \tilde{\mathcal{A}}(\tilde{x}) \to \Pi$ such that for any $\tilde{a} \in \tilde{\mathcal{A}}(\tilde{x})$, if $\Psi(\tilde{a}) = (I, \pi, \psi)$ then $S(\tilde{x}) \subset I$.[1]

Henceforth we will use "abstraction" instead of "option-based abstraction" and will call $(\tilde{\mathcal{X}}, \tilde{\mathcal{A}}, \tilde{p}, \tilde{c})$ the "abstract MDP", $\tilde{\mathcal{X}}$ the set of abstract states, $\tilde{\mathcal{A}}$ the set of abstract actions, etc. Notationally, we call $(X, A, p, c)$ the *ground level* MDP, and we will identify quantities related to the abstract MDP by using a tilde ( $\tilde{}$ ). For simplicity, we will identify the abstract actions with their corresponding options. In particular, we will call $\tilde{a}$ both an abstract action and an option, depending on the context.

In the following, we will assume that $\{S(\tilde{x}) \,|\, \tilde{x} \in \tilde{\mathcal{X}}\}$ is a partition of $X$; we can then let $\tilde{x} : X \to \tilde{\mathcal{X}}$ denote the (unique) abstract state that includes $x$: $\tilde{x}(x) \in \tilde{\mathcal{X}}$ such that $x \in S(\tilde{x}(x))$, and say that $(\tilde{\mathcal{X}}, S)$ is an *aggregation of the states* in $X$. We also define $S(x) = S(\tilde{x}(x))$ as the set of states in $X$ that are in the same partition with $x$.

The restriction on $\Psi$ in the above definition ensures that the execution of any policy $\tilde{\pi}$ in the abstract MDP is well-defined and proceeds as follows. Initially, there is no active option. In general, whenever there is no active option, we look up the abstract state $\tilde{x} = \tilde{x}(x)$ based on the current state $x$ and activate the option $\Psi(\tilde{\pi}(\tilde{x}))$. When there is an active option, the option remains active until the corresponding terminating condition fires. When an option is active, the option's policy selects the actions in the ground level MDP. This way a policy $\tilde{\pi}$ in the abstract MDP *induces* a policy in the ground level MDP.

Our goal now is to characterize what makes an abstraction accurate. The following theoretical analysis is novel as it considers abstractions where the action set is changed. In particular, the action set can potentially be reduced and the abstract actions can be options. To our knowledge, such options-based abstractions have not been analyzed previously; the closest results are probably Theorem 2 of Kim and Dean (2003) and Theorem 4 of Dean, Givan, and Leach (1997). The proof is rather technical and is given in the extended version of our paper (Isaza, Szepesvári, Bulitko, & Greiner, 2008).

Consider a proper policy $\pi$ of the ground level MDP. We want abstractions such that one can always find a policy in the abstract MDP $(\tilde{\mathcal{X}}, \tilde{\mathcal{A}}, \tilde{p}, \tilde{c})$ that approximates $\pi$ well, no matter how $\pi$ was chosen. Clearly, this depends on how the action set $\tilde{A}$ and the corresponding transitions and costs are defined in the abstract MDP. Quantifying this requires a few definitions: Let $\tilde{p}_\pi(\tilde{x}, \tilde{y})$ be the probability of landing in some state of $S(\tilde{y})$ when following policy $\pi$ until it leaves the states of $S(\tilde{x})$, when the initial state is selected at random from the states of $S(\tilde{x})$ based on the distribution $\mu_{S(\tilde{x})}$. Let $\tilde{c}_\pi(\tilde{x})$ denote the corresponding expected "immediate" cost. Now pick a proper policy $\tilde{\pi}$ of the abstract MDP. Let $\tilde{w}$ be the weight vector that makes $T_{\tilde{\pi}}$ a contraction in the abstract MDP. Further, define $\tilde{p}_{\tilde{\pi}}(\tilde{x}, \tilde{y}) = \tilde{p}(\tilde{y}|\tilde{x}, \tilde{\pi}(\tilde{x}))$ and $\tilde{c}_{\tilde{\pi}}(\tilde{x}) = \sum_{\tilde{y}\in\tilde{\mathcal{X}}} \tilde{p}_{\tilde{\pi}}(\tilde{x}, \tilde{y}) \tilde{c}_{\tilde{\pi}}(\tilde{x}, \tilde{y})$ and the mixed $\ell^1/\ell^\infty$ norm $\|\cdot\|_{\tilde{w},1/\infty}$:

$$\|\tilde{p}_1 - \tilde{p}_2\|_{\tilde{w},1/\infty} = \max_{\tilde{x}\in\tilde{\mathcal{X}}} \sum_{\tilde{y}\in\tilde{\mathcal{X}}} |\tilde{p}_1(\tilde{x}, \tilde{y}) - \tilde{p}_2(\tilde{x}, \tilde{y})| \frac{\tilde{w}(\tilde{y})}{\tilde{w}(\tilde{x})}.$$

Let

$$\varepsilon_{\pi,\tilde{\pi}} = \|\tilde{c}_\pi - \tilde{c}_{\tilde{\pi}}\|_{\tilde{w},\infty} + c_{\max} \|\tilde{p}_\pi - \tilde{p}_{\tilde{\pi}}\|_{\tilde{w},1/\infty}, \qquad (1)$$

where $c_{\max}$ is the maximum of the immediate costs in the ground level MDP. Hence, $\varepsilon_{\pi,\tilde{\pi}}$ measures how well the costs and the transition probabilities induced by $\pi$ "after state aggregation" match those of $\tilde{\pi}$. Introduce $c(x, \pi)$ as the expected total cost incurred, conditioned on that policy $\pi$ starting in state $x$ and stopping when it exits $S(x)$. Further, introduce $p(\tilde{y}|x, \pi)$ as the probability that, given that policy $\pi$ is started in state $x$, when it exits $S(x)$ it enters $S(\tilde{y})$ ($\tilde{y} \neq \tilde{x}(x)$). Now fix an abstract state $\tilde{x} \in \tilde{\mathcal{X}}$. If the costs $\{c(x, \pi)\}_{x\in S(\tilde{x})}$ and probabilities $\{p(\tilde{y}|x, \pi)\}_{x\in S(\tilde{x})}$, $\tilde{y} \neq \tilde{x}$, have a small range then we can model closely the behavior of $\pi$ locally at $S(\tilde{x})$ by introducing an option with initial states in $S(\tilde{x})$ which mimics the "expected" behavior of $\pi$ as it leaves $S(\tilde{x})$, assuming, say, that the initial state in $S(\tilde{x})$ is selected at random according to the distribution $\mu_{S(\tilde{x})}$. If we do so for all abstract states $\tilde{x} \in \tilde{\mathcal{X}}$ then we can make sure that $\min_\pi \varepsilon_{\pi,\tilde{\pi}}$ is small. If the above range conditions hold for all policies $\pi$ of the ground level MDP and all abstract states $\tilde{x} \in \tilde{\mathcal{X}}$ then by introducing a sufficiently large number of abstract actions it is possible to keep $\max_\pi \min_{\tilde{\pi}} \varepsilon_{\pi,\tilde{\pi}}$ small. Further, notice that $\max_\pi p(\tilde{y}|x, \pi)$ is zero unless there exists a transition from some state of $S(x)$ to some state of $S(\tilde{y})$, in which case we say that $S(x)$ is connected to $S(\tilde{y})$. Hence, no abstract action is needed "between" $\tilde{x}$ and $\tilde{y}$, unless $S(\tilde{x})$ is connected in the ground level MDP to $S(\tilde{y})$.

Define $\tilde{T}_\pi : B(\tilde{X}) \to B(\tilde{X})$, $(\tilde{T}_\pi \tilde{v})(\tilde{x}) = \tilde{c}_\pi(\tilde{x}) + \sum_{\tilde{y}} \tilde{p}_\pi(\tilde{x}, \tilde{y}) \tilde{v}(\tilde{y})$. Since $\pi$ is proper in the ground level MDP, it is not difficult to show that $\tilde{T}_\pi$ is a contraction with respect to an appropriately defined weighted supremum norm.

The next result gives a bound on the difference of value functions of $\pi$ and $\tilde{\pi}$ in terms of $\varepsilon_{\pi,\tilde{\pi}}$:

[1] $2^X$ denotes the power set of $X$: the set of all subsets of $X$.

**Theorem 1** Let $\pi$ be a proper policy in the ground level MDP and let $\tilde{\pi}$ be a proper policy in the abstract MDP. Let $w_\pi$ (resp., $\tilde{w}_\pi$) be the weight vector that makes $T_\pi$ (resp., $\tilde{T}_\pi$) a contraction and let the corresponding contraction factor be $\gamma_\pi$ (resp., $\tilde{\gamma}_\pi$). Let $v_\pi$ be the value function of $\pi$ and $\tilde{v}_{\tilde{\pi}}$ be the value function of $\tilde{\pi}$. Then

$$\|v_\pi - E\tilde{v}_{\tilde{\pi}}\|_{w,\infty} \le \frac{\|Av_\pi - v_\pi\|_{w,\infty}}{1-\gamma_\pi} + \lambda_\pi \frac{\varepsilon_{\pi,\tilde{\pi}}}{1-\tilde{\gamma}_\pi},$$

where the operator $E$ extends functions defined over $\tilde{\mathcal{X}}$ to functions defined over $X$ in a piecewise constant manner: $E : B(\tilde{\mathcal{X}}) \to B(X)$, $(E\tilde{v})(x) = \tilde{v}(\tilde{x}(x))$, and $A : B(X) \to B(X)$ is the aggregation operator defined by

$$(AV)(x) = \sum_{z \in S(x)} \mu_{S(x)}(z) V(z),$$

and $\lambda_\pi = \max_{x \in X} \tilde{w}_\pi(\tilde{x}(x))/w_\pi(x)$.

The factor $\lambda_\pi$ measures how many more steps are needed to reach the goal if the execution of policy $\pi$ is modified such that, whenever the policy enters a new cluster $\tilde{x}$, the state gets perturbed, by choosing a random state according to $\mu_{S(\tilde{x})}$.

The theorem provides a bound on the difference between the value function of a ground-level policy $\pi$ and the value function of an abstract policy when its value function is extended to the ground-level states. The bound has two terms: The first bounds the loss due to state abstraction, while the second bounds the loss due to action abstraction. When a similar range condition holds for the abstract actions, too, then it is possible to bound the difference between the value function of the policy induced in the ground level MDP by $\tilde{\pi}$ and $E\tilde{v}_{\tilde{\pi}}$, yielding a difference on the value functions of $\pi$ and the policy induced by $\tilde{\pi}$. Isaza et al. (2008) provides further details.

If we apply this result to an optimal policy $\pi^*$ of the ground level MDP, we immediately get a bound on the quality of the abstraction. We may conclude then that the quality of abstraction is determined by the following factors: (i) whether states with different optimal values are aggregated; (ii) whether the random perturbation described in the previous paragraph can increase the number of steps to the goal substantially; and (iii) whether the immediate costs $\tilde{c}_{\pi^*}$ and transition probabilities $\tilde{p}_{\pi^*}$ can be matched in the abstract MDP.

Since we want to build abstractions that work independently of where the goal is placed, the knowledge of the optimal policy with respect to a particular goal cannot be exploited when constructing the abstractions. In order to prevent large errors due to (i) and (ii), we restrict aggregation such that only a few states are grouped together. This makes the job of creating an aggregation easier. Fortunately, we can achieve higher compression by adding additional layers of abstractions. We can address (iii) by creating a sufficiently large number of abstract actions. Here, we use the simplifying assumption that we only create abstract actions that bring the agent from some cluster of states to some neighboring cluster. These can serve as a "basis" for matching any complex next-state distribution over the clusters by choosing an appropriate stochastic policy in the abstract MDP. We also want to ensure that the initial state within a cluster has a small influence on the probability of transitioning to some neighboring cluster and the associated costs. We use two constants, $\varepsilon$ and $\mu$, to bound the amount of variation with respect to initial states; note this allows us to control the difference between the value function of a policy induced in the ground level MDP by some abstract policy $\tilde{\pi}$ and the extension of the value function of $\tilde{\pi}$ defined in the abstract MDP to the ground level states, $E\tilde{v}_{\tilde{\pi}}$. This is necessary to ensure that a good policy in the abstract MDP produces a good policy in the ground-level MDP, ultimately assuring that the optimal policy of the abstract MDP will give rise to a close to optimal policy in the ground-level MDP. The resulting procedure is described in the next section.

## 3 Abstracting an SSP

This section describes our algorithm **BuildAbstraction** for automatically building options-based abstractions. These abstractions are goal-independent and thus apply to a series of SSPs that share the state space and transition dynamics. The process consists of four main steps (Figure 1): (1) *Cluster* proposes candidates for abstract states; (2) *GenerateLinkCandidates* proposes candidates for abstract actions (or "links"); (3) *Repair* validates and, if necessary, repairs the links in order to satisfy the so-called $(\varepsilon, \mu)$-connectivity property (the formal definition is given later) and *Prune* discards excessive links.

Once an abstraction is built, we use a special-purpose planning procedure (described in Section 4) to solve specific SSPs. The rest of this section describes the four steps of our **BuildAbstraction** algorithm in detail.

**Step 1:** *Cluster.* A straightforward cluster-er will cluster a state with some of its immediate neighbors. Unfortunately, this approach may group states with diverging trajectories (the trajectories from one state can differ from those of the other state). By looking for the peers of a state (predecessors of its successors, line 2, Figure 1) we hope to find a peer whose trajectories are similar to the trajectories of the first state. Note that the clustering routine creates minimal clusters. This is advantageous as it means the subsequent steps, which connect clusters, is more likely to succeed. Unfortunately, it also means relatively low reduction in the number of states. Several layers of abstractions can help increase this reduction.

**Step 2:** *Generate Link Candidates.* After forming the initial clusters (i.e., the initial abstract states), **BuildAbstraction** generates candidates for abstract actions. One approach is simply to propose abstract actions for *all* pairs of abstract states, in the hope that only important ones will remain after pruning. We use a less expensive strategy and propose abstract action candidates only for "nearby" clusters (line 8). For each such pair we add two candidate links: one in the forward and another in the backward direction — this heuristic quickly generates reasonable link candidates. We typically use $k = 1$. Our experiments confirm this is

**BuildAbstraction**$(k, p, M)$ // $M$ – ground level MDP
*–Cluster–*
1 **for** each unmarked ground state $x$ **do**
2 Find $P(x)$, all the predecessors of successors of $x$
3 Find $y \in P(x)$ that has the most successors in common with $x$
4 Add $\tilde{x}$ to $\tilde{X}$ with $S(\tilde{x}) = \{x, y\}$
5 Mark states $\{x, y\}$
6 **end for**
*–GenerateLinkCandidates–*
7 repairQ $\leftarrow \emptyset$
8 **for** every $\tilde{x}, \tilde{y} \in \tilde{X}$, where any state in $S(\tilde{y})$ is within $k$ ground transitions of some state in $S(\tilde{x})$ **do**
9 repairQ $\leftarrow$ repairQ $\cup \{(\tilde{x}, \tilde{y}),\ (\tilde{y}, \tilde{x})\}$
10 **end for**
*–Repair–*
11 **while** repairQ $\neq \emptyset$ **do**
12 $(\tilde{x}, \tilde{y}) \leftarrow$ pop an element from repairQ
13 set up an SSP, $S$, with domain $R \subset X$ where $S(\tilde{x}) \cup S(\tilde{y}) \subset R$ with states in $S(\tilde{y})$ as goals
14 attempt to find an optimal policy $\pi_S$ in $S$ with IPS
15 **if** no policy found **then**
16 **continue**
17 **else if** $\pi_S$ does not meet the $(\varepsilon, \mu)$ conditions **then**
18 split the cluster adding both parts to repairQ
19 **else**
20 add $\tilde{a}$ to $\tilde{A}(\tilde{x})$ with $\Psi(\tilde{a}) = (S(\tilde{x}), \pi_S, \mathbb{I}_{S(\tilde{y})})$ [2]
21 set $\tilde{c}(\tilde{x}, \tilde{a})$ to be the expected cost of executing $\tilde{a}$ from a random state of $\tilde{x}$
22 set $\tilde{p}(\tilde{y}|\tilde{x}, \tilde{a}) = 1$, $\tilde{p}(\tilde{y}'|\tilde{x}, \tilde{a}) = 0$ for $\tilde{y}' \neq \tilde{y}$.
23 **end if**
24 **end while**
*–Prune–*
25 **for** each state $\tilde{x}$ **do**
26 find $\tilde{A}^*(\tilde{x}) = \{\tilde{a}_1, \dots, \tilde{a}_m\}$, all abstract actions that connect clusters that are neighbors in $M$
27 order $\tilde{A}(\tilde{x}) \setminus \tilde{A}^*(\tilde{x})$ to create $[\tilde{a}_{m+1}, \dots, \tilde{a}_n]$ such that $\tilde{c}(\tilde{x}, \tilde{a}_i) \leq \tilde{c}(\tilde{x}, \tilde{a}_{i+1})$, $i = m+1, \dots, n-1$
28 let $\tilde{A}(\tilde{x}) = \{\tilde{a}_1, \dots, \tilde{a}_p\}$
29 **end for**
30 **return** $(\tilde{X}, \tilde{A}, \tilde{p}, \tilde{c})$.

Figure 1: The abstraction algorithm.

sufficient; increasing $k$ results in slightly better quality, but slower running times when solving the planning problems.

**Step 3:** *Repair.* For each candidate abstract action connecting abstract states $\tilde{x}$ and $\tilde{y}$, we first need to derive an option that, starting in any state in cluster $\tilde{x}$ leads the agent to some state in cluster $\tilde{y}$ with a minimum total expected cost. We derive this option by setting up a shortest path problem $S$, whose domain includes $S(\tilde{x})$ and $S(\tilde{y})$. We set the domain of $S$ to be sufficiently large that a policy within this domain can reliably take the agent from any state of $S(\tilde{x})$ to some state of $S(\tilde{y})$. **BuildAbstraction** builds this domain by performing a breadth-first search from $S(\tilde{y})$, proceeding backwards along the transitions, stopping at depth $D + m$, where $D$ is the search depth from $S(\tilde{y})$ and $m$ is the margin to leave after all states of $S(\tilde{x})$ were added to the domain. If there is any state of $S(\tilde{x})$ that was not included at depth $D$, the *Repair* routine reports 'no solution'. The transitions, actions and costs of $S$ are inherited from the MDP $M$. We also add a new terminating state, which is the destination of transitions leaving the region — i.e., those transitions are redirected to this new terminal, with a transition cost that exceeds the maximum of the total expected costs of the ground level MDP. The high cost discourages the solutions to enter the extra terminating state. The optimal solution to $S$ is obtained by using the *Improved Prioritized Sweeping* (IPS) algorithm of McMahan and Gordon (2005), (line 14). We selected this algorithm based on its excellent performance and known optimality properties (IPS reduces to Dijkstra's method in deterministic problems). The resulting policy $\pi$ is checked against $(\varepsilon, \mu)$-connectivity, defined as follows: we first compute the expected total cost of reaching some states in $S(\tilde{y})$ for all states of $S(\tilde{x})$; let the resulting costs be $c(x, \pi)$. Similarly, we compute the probabilities $p(S(\tilde{y})|x, \pi)$ for every $x \in S(\tilde{x})$. Then we check if $\max_{x,x' \in S(\tilde{x})} |c(x, \pi) - c(x', \pi)| \leq \varepsilon$ and $\max_{x,x' \in S(\tilde{x})} |p(S(\tilde{y})|x, \pi) - p(S(\tilde{y})|x', \pi)| \leq \mu$ both hold. If these constraints are met, a new abstract action is created and is added to the set of admissible actions at $\tilde{x}$ and the policy is stored as the option corresponding to this new abstract action (lines 20–22). Otherwise, the cluster is split (since every cluster has two states, this is trivial) and the appropriate link candidates are added to the repair queue so that no link between potentially connected clusters is missed.

**Step 4:** *Prune.* After step 3, we have an abstract SSP whose abstract states are $(\varepsilon, \mu)$-connected. However, our abstract action generation mechanism may produce too many actions, which may slow down the planning algorithm (see Section 4). We address this problem using a pruning step that leaves only the "critical" and cheapest abstract actions. An action is "critical" if it connects clusters that are connected at the ground level with a single transition; these actions are important to keep the structure of the ground level MDP. We also keep the cheapest abstract actions as they are likely to help achieve high quality solutions. The "pruning parameter", $p$, specifies the total number of actions to keep. (If $p$ is smaller or equal than the number of ground actions, then only the "critical" actions are kept.)

**BuildAbstraction** runs in time linear in the size of the input MDP, as every step is restricted to some fixed size neighborhood of some state (i.e., every step is local). Further, employing a suitable data structure, the memory requirements can also be kept linear in the size of the input. These properties are important when scaling up to realistic, real-world problem sizes.

# 4 Planning with an Abstraction

After building an abstraction, we can use it to solve particular SSP problems. When we specify a new goal, our abstraction planner, **AbsPlanner**, then creates a *goal-approach region* in the abstract MDP that includes the goal and is large enough to include all states of the cluster containing the goal. **AbsPlanner** builds this region by starting with the ground goal and adding states and transitions in a breadth-first fashion to a certain depth, proceeding backwards along the transitions, stopping only after adding all states of the goal-cluster. After building the region, **AbsPlanner** produces an SSP. The domain of this

[2]Here $\mathbb{I}_S$ is the characteristic function of $S$: $\mathbb{I}_S(x) = 1$ iff $x \in S$ and $\mathbb{I}_S(x) = 0$ otherwise.

SSP includes the states found in the breadth-first search, and also a new terminal state that becomes the destination of transitions leaving the region — i.e., those transitions are redirected to this new terminal, with a high transition cost. All other costs and transitions of this SSP are inherited from the ground level MDP. **AbsPlanner** uses IPS to solve the local MDP, and saves the resulting *goal-approach policy*. It then solves the abstract MDP, where the goal cluster is set as the goal. When executing the resulting policy $\tilde{\pi}$, **AbsPlanner** proceeds normally until reaching a state of the goal-approach region; it then switches to the goal-approach policy, which it follows until reaching the goal or leaving the region. When this latter event happens and the state is $x$, execution switches to the option $\tilde{\pi}(\tilde{x}(x))$.

When using multiple levels of abstraction, **AbsPlanner**'s execution follows a recursive, hierarchical strategy. Note that the size of the goal-approach region is independent of the size of the MDP. Thus, the planning time will depend on the size of the top-level abstract MDP. For an MDP of size $n$, by using $\log n$ levels of hierarchy, in theory it is then possible to achieve planning times that scale with $O(\log n)$. However, depending on the problem, it might be hard to guarantee high quality solutions when using many levels of abstraction. Furthermore, in practice (over the problems used in our tests), the computation time is dominated by the time needed to set up and solve the goal-approach SSPs, which is required for even one layer of abstraction. This is partly because our abstractions result in *deterministic* shortest path problems, whose solutions can be found significantly faster than those of stochastic problems.

## 5 Empirical Evaluation

This section summarizes our empirical evaluation of this approach, in terms of the quality (suboptimality) of the solutions and the solution times. Here we report the trade-offs of using different levels of abstraction as well as the dependence on the "stochasticity" of the transitions. (Note that stochasticity makes it difficult to build abstractions.) We also tested the performance of the algorithm on more practical problems. In addition to the results presented here, we conducted extensive experiments, studying the trade off between solution quality and solution time as a function of the various parameters of our algorithm (e.g., the values of $p$, $k$, or the number of abstraction levels), the scaling behavior of our algorithm in terms of its resource usage, the quality of solutions and the solution time. These results, appearing in (Isaza et al., 2008), confirm that the algorithm is robust to the choices of its parameters and scales as expected by increasing problem sizes.

We run experiments over three domains: noisy gridworlds, a "river" and congested game maps. The gridworlds are empty and have four actions: *up*, *down*, *left*, and *right*, each with cost 1. The probability that an action leaded to the expected position (e.g., the action *up* moves the agent up one cell) is $0.7$, while the probability of reaching any of the other three adjacent cells is $0.1$.

The river is similar to the gridworld: its dimensions are $w \times h$, but there is a current flowing from left to right and a fork corresponding to a line connecting the points $(w/2, h/2)$ and $(w, h/2)$.[3] The flow is represented by modifying both the cost structure and the transition probabilities of the actions: action *forward* costs $1$, *backward* costs $5$, *diagonally-up-and-forward* and *diagonally-down-and-forward* each cost $\sqrt{2}$. These actions are also stochastic: For the *backward* action, the probabilities are $0.7$ for going back and $0.1$ for each of the other actions. For the other three actions, the anticipated move occurs with probability $0.6$ and the other moves except *backwards* occur each with probability $0.2$, and *backwards* has probability $0$. We include the river domain to determine whether our system can deal with non-uniform structures and because the fork complicates the task of creating abstractions. We empirically found the time to build abstractions for the $n$-state gridworld was close to $n/100$ seconds, and around $n/50$ for an $n$-state river domain. The build time for the maps, using $k = 1$, was between 75 and 100 seconds.[4]

The congested game maps are again similar to gridworlds, but with obstacles and with transitions probabilities that depend on the congestion. The obstacle layout comes from commercial game maps, and the stochastic dynamics simulate what happens if multiple units traverse the same map: in narrow passages, the units to become congested, which means an agent trying to traverse such a passage is likely to be blocked. We model this by modifying each action by including a probability that the action will "fail" and cause the agent to stay at the same position. This "failure probability" depends on the position on the game map, calculated by simulating many units randomly traversing the game maps and measuring the average occupation of the individual cells, then turning the occupation numbers into probabilities. The optimal policy of an agent in a congested game map will then try to avoid narrow passages, since the higher probability of traffic congestion in such regions means an agent takes much longer to get through those regions.

The baseline performance measures are obtained by running the state-of-the-art SSP solver algorithm IPS. For each study, we generate the abstraction and then use it to solve 1,000 problems, whose start and goal locations are selected uniformly at random. For each problem we measure the solution time in seconds and the total solution cost for both IPS and our method, then compute the geometric average of the individual suboptimalities and the individual solution time ratios.

### 5.1 Abstraction level trade-offs

We used a $100 \times 100$ gridworld to analyze the trade-offs of different abstraction levels, with several different parameter configurations. We say a configuration is "dominant" if it was a Pareto optimal — i.e., if no other configuration is better in both time and suboptimality.

Figure 2 presents properties of the dominant configurations for various abstraction levels. We see that using a smaller number of abstractions required more time but produced better solutions (i.e., lower suboptimality), and higher levels of abstractions required less solution time but produced inferior solutions (i.e., increased suboptimality). Note that there are dominant configurations for every level of abstraction, from 0 to 5.

[3] See (Isaza et al., 2008) for more details, including relevant pictures.

[4] We ran all experiments on a 2GHz AMD Opteron(tm) Processor with 4GB of RAM running Linux with kernel 2.6.18.

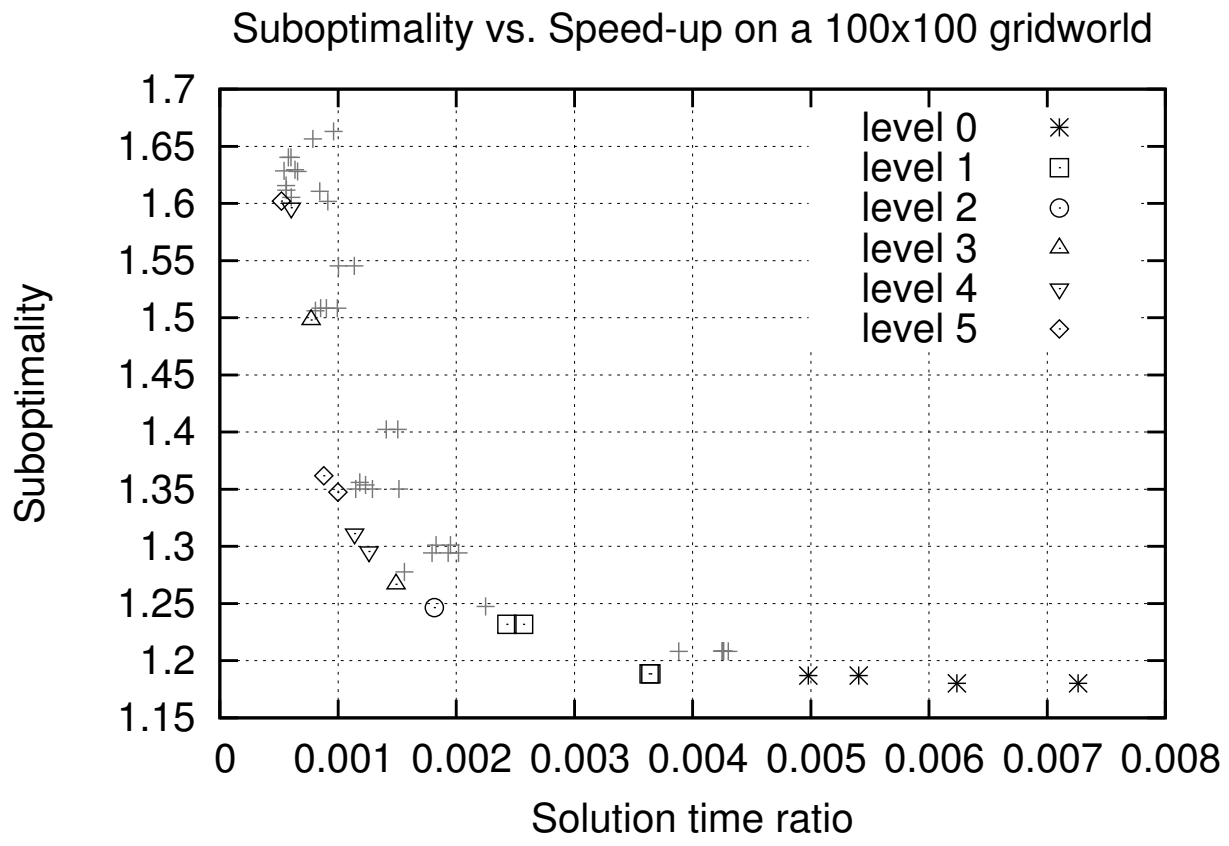

Figure 2: Subobtimality versus the solution time ratio as compared to IPS for different parameter configurations. The dominant configurations are shown for different levels of abstraction.

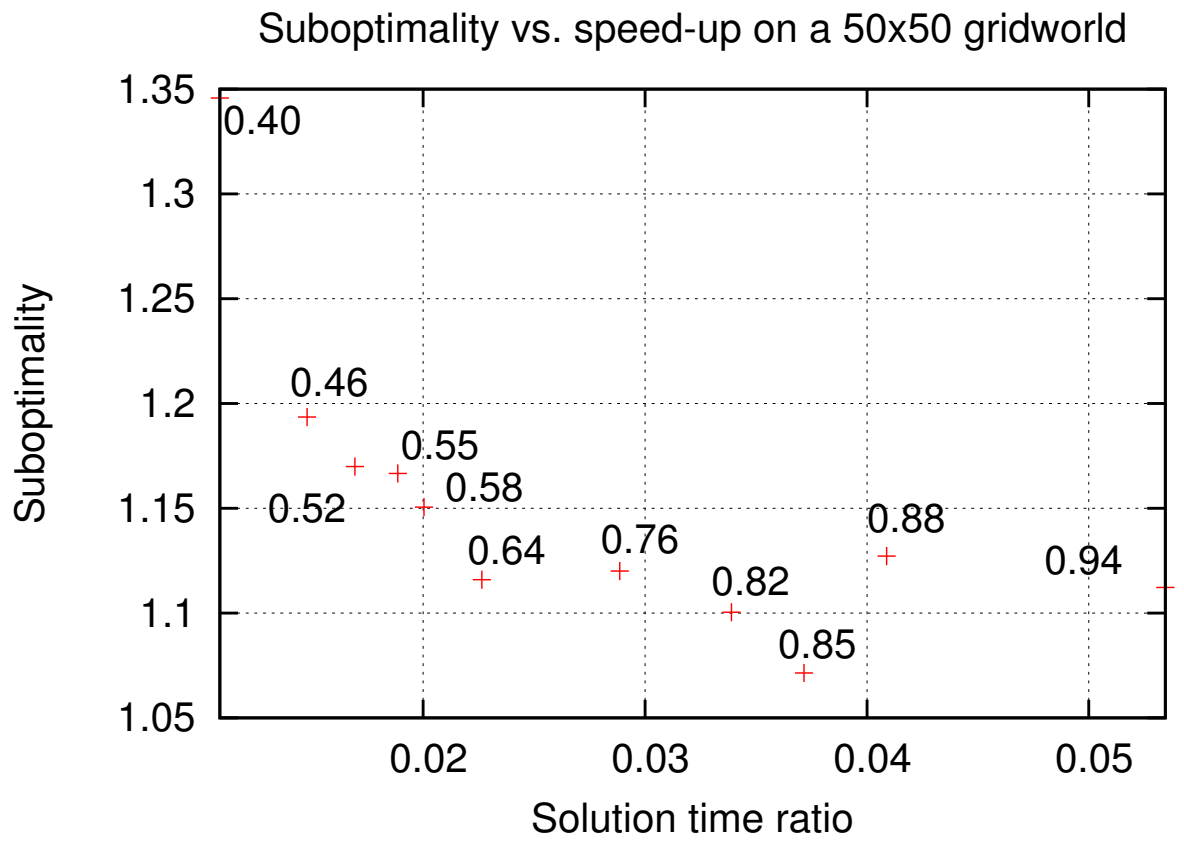

Figure 3: Subobtimality versus the solution time ratio as compared to IPS for different values of $P$.

We obtain a "level 0" abstraction by converting the given ground-level SSP to *deterministic* shortest path problem with the same states. (Recall that our abstraction process abstracts the state space and produces a *deterministic* SSP; here we just used the original state space.) Figure 2 shows that this transformation provides solutions whose quality is slightly inferior to the original problem, but it finds this solution significantly faster (e.g., in 0.005 to 0.0073 of the time). We also see that these "level 0" solutions are superior to those based on higher abstraction levels, but one can obtain these level-$i$ solutions in yet less time.

## 5.2 Sensitivity to Stochasticity of the Dynamics

As the environment becomes noisier, it becomes more difficult to construct a high quality abstraction. This section quantifies how the solution quality and construction time relate to noise in the dynamics. In general, we consider an action "successful" if the agent moves to the appropriate direction; our gridworld model set the success probability to $P = 0.7$, leaving a probability of $(1 - P)/3$ to moving in each of the other three directions. Here, we vary the value of $P$. All of these experiments use a $50 \times 50$ gridworld with $k = 2$ and $p = 4$ (which means we keep only the "critical" actions; see Section 3).

Figure 3 plots the suboptimality and the speed-up of finding a solution using our method, as compared to IPS, for different values of $P$. We see that our method loses optimality as the dynamics becomes noisier (i.e., when $P$ gets smaller). This is because our abstract actions, trying to move the agent from one abstract state to the next will fail with higher probability for noisier dynamics. Note that the advantage of our method, in terms of planning time, becomes larger with increased stochasticity. This is because our abstractions are deterministic and planning in a deterministic system is much faster than planning with a stochastic system.

Figure 4, plotting the absolute values of cost and time for both our method and IPS, provides another insight: It shows that for increasing stochasticity both methods are slowed down, but our method can cope better with this situation. This figure also confirms that this leads to a loss in solution quality.

For our method the typical parameters produce a suboptimality of around 1.4 for the river, and around 1.25 for the gridworld domain. The speed-up for the gridworld is around 30, while for the river it is around 800.

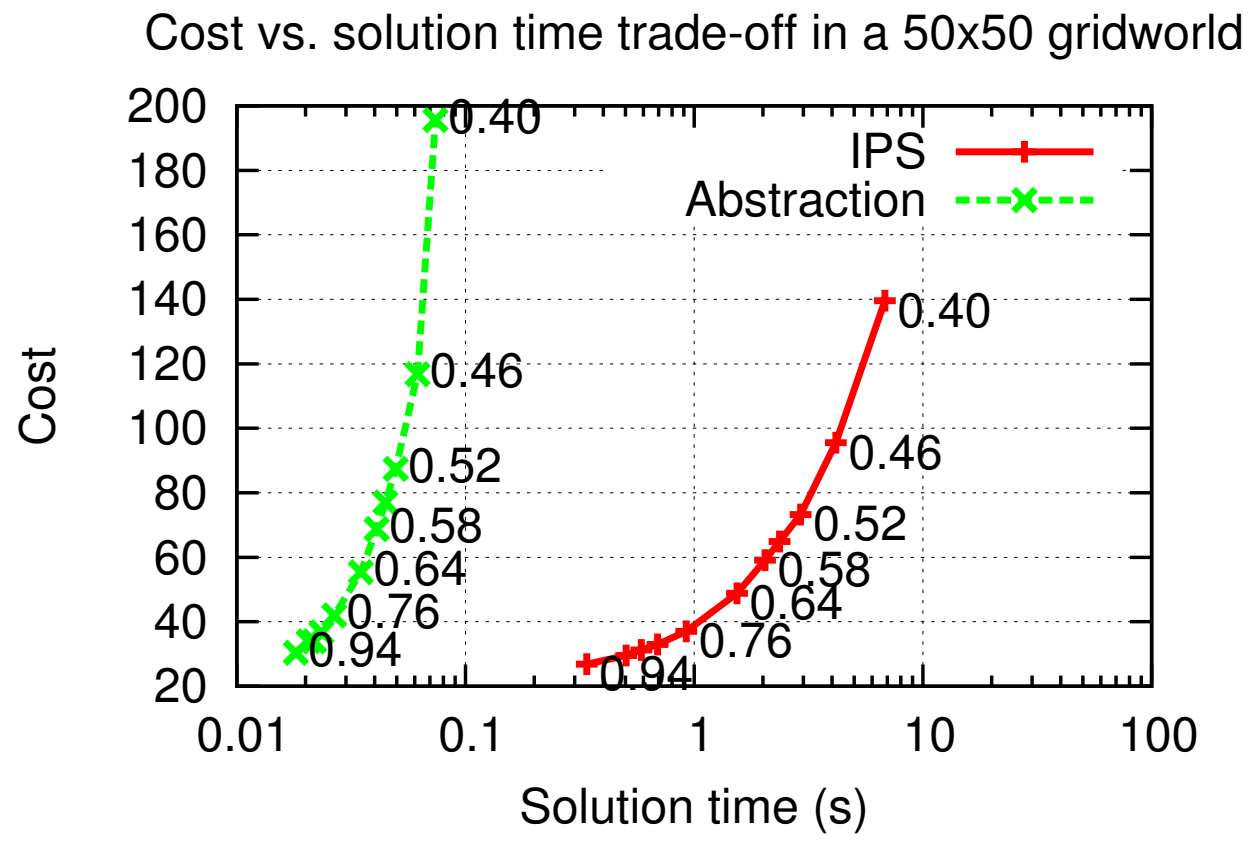

Figure 4: Cost versus solution time for IPS and abstraction at different values of $P$.

## 5.3 Congested Game Maps

To test the performance of our approach in a more practical application, we used maps modeled after game environments from commercial video games. We first created simplified gridworlds that resemble some levels from a popular role-playing and real-time-strategy game. We then converted the gridworlds into congested maps as described earlier. This produced maps with state space sizes of 6176 (BG1), 5672 (BG2), 5852 (BG3), 20249 (WC1) and 9848 (WC2). Figure 5 provides one such map, where each state's color indicates the associated congestion: warmer/redder colors indicates high congestion (i.e., low probability of success $P$) while colder/bluer colors indicates low congestion (i.e., high value of $P$). Very dark blue indicates impassable obstacles. We see that many of the states in cluttered regions are highly congested and should therefore be avoided.

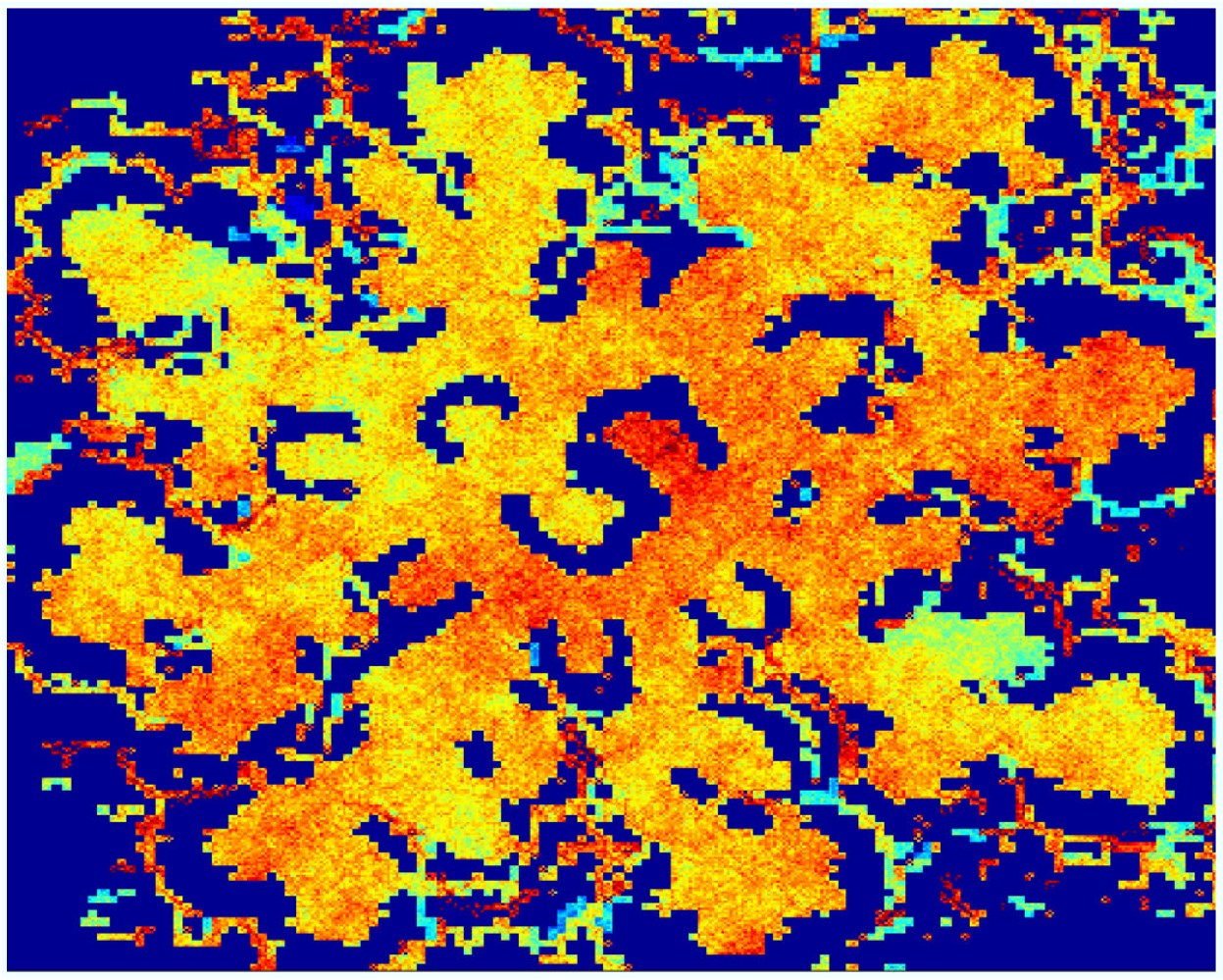

Figure 5: A congested game map. Darker/redder color refers to high congestion. Dark blue regions are impassable obstacles.

Figure 6 shows the solution time and the solution suboptimalities for both our method and IPS, for two maps from WarCraft (Blizzard Entertainment, 2002) and three maps from Baldur's Gate (BioWare Corp., 1998), including Figure 5, using only a single layer of abstraction. We see that our approach is indeed successful in speeding up the planning process, while keeping the quality of the resulting solutions high.

## 6 Related Work

Due to space constraints we review only the most relevant work; references to other related works can be found in the extensive bibliography lists of the cited works. Dean et al. (1997) introduced the notion of $\varepsilon$-homogeneous partitions and analyzed its properties, but without giving explicit loss bounds. Kim and Dean (2003) developed some loss bounds. Their Theorem 2 can be strengthened with our proof method to $\|v^* - v_P^*\|_\infty \leq \|Tv_P^* - v_P^*\|_\infty / (1-\gamma)$ (using our notation), basically dropping the first term in their bound. Here $v^*$ is the optimal value function in the original MDP, $v_P^*$ is the optimal value function of the aggregated MDP extended back to the original state space in a piecewise constant manner and $T$ is the Bellman-optimality operator in the original MDP. This bound is problematic as it does not show how the quality of partitions influences the loss. Our bound improves on this bound in this respect, and also by extending it to the case when the abstract actions correspond to options. While Asadi and Huber (2004) also considered such options-based abstractions, they assume that the abstract actions (options) are given externally (possibly by specifying goal states for each of them) and they do not develop bounds. In a number of subsequent papers, the authors refined their methods. In particular, they became increasingly focussed on learning problems. For example, in the recent follow-up work, Asadi and Huber (2007) provide a method to learn an abstract hierarchical representation that uses state aggregation and options. Since they are interested in skill transfer through a series of related problems that can *differ* in their cost structure, they introduce a heuristic to discover subgoals based on bottleneck states. They learn options for achieving the discovered subgoals and introduce a partitioning that respects the learned options (in the clusters typically there are many states). The success of the approach relies critically on the existence of meaningful bottleneck states. This leads to a granularity issue: identifying the bottleneck states requires computing a statistic for each state visited, meaning bottlenecks will not be pronounced if resolution is increased in narrow pathways. Nevertheless, the approach has been successfully tested in a non-trivial domain of 20,000 states.

Hauskrecht, Meuleau, Kaelbling, Dean, and Boutilier (1998) introduce a method that also uses options, but the abstract states correspond to boundary states of regions. The regions are assumed to be given *a priori*. The idea is similar to using bottleneck states. In contrast to that work, we do not assume any prior knowledge, but construct the abstractions completely autonomously. Further, we deal with undiscounted SSPs, while Hauskrecht et al. (1998) dealt with discounted MDPs (but this difference is probably not crucial).

## 7 Discussion and Future Directions

In the approach presented, options serve as closed-loop abstract actions. Another way to use an abstract solution would be to use the abstract value function to guide local search initiated from the current state. These ideas has proven successful in pattern-database research where the cost of an optimal solution of an abstract problem is used as a powerful heuristic for the original problem. Such a procedure has the potential to improve solution quality, while keeping low the cost of the planning steps interleaved with execution. Another idea is to use the abstraction to select the amount of such local search (i.e., the depth of the rollouts); these ideas has proven successful in deterministic environments (Bulitko, Björnsson, Luštrek, Schaeffer, & Sigmundarson, 2007; Bulitko, Luštrek, Schaeffer, Björnsson, & Sigmundarson, 2008).

Presently, our abstractions are deterministic. This suggests two avenues for future work. First, applying advanced heuristic search methods to such abstractions may lead to performance gains. Second, in highly stochastic domains, the abstraction's determinism may lead to a poor quality of solution, as the cost of ensuring arrival at an abstract state with certainty (or very high probability) can lead to very conservative and costly paths. Thus, it would be of interest to investigate *stochastic* abstractions. One idea is to modify the way abstract actions are defined: When planning to connect to abstract states after a solution of the local SSP is found, with a little extra work we can compute the probabilities of reaching various neighboring abstract states under the policy found when the policy leaves the region of interest.

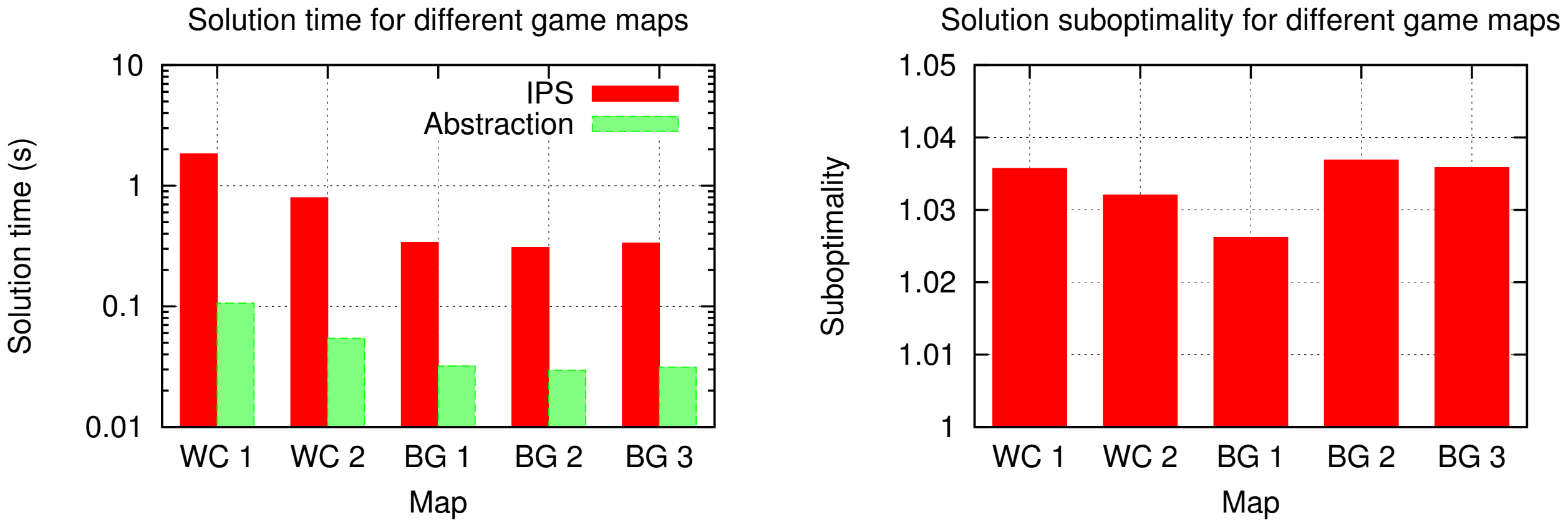

Figure 6: Solution times (left) and suboptimalities (right) for several game maps.

Yet another avenue for future work would be to move from a state-based problem formulation to a feature-based one, assuming that the features describe the states. The challenge is to design an algorithm that can construct an abstraction without enumerating all the states, as ours currently does. Although this paper has not attempted to address this problem, we believe that the approach proposed here (i.e., incremental clustering and defining options by solving local planning problems) is applicable.

Finally, although the present paper dealt only with undiscounted, stochastic shortest path problems, the approach can be extended to work for *discounted* problems. This holds because a discounted problem can always be viewed as an undiscounted stochastic shortest path problem where every time step a transition is made to some terminal state with probability $1-\gamma$, where $0<\gamma<1$ is the discount factor.

## 8 Conclusions

This paper has explored ways to speed up planning in SSP problems via goal-independent state and action abstraction. We strengthen existing theoretical results, then provide an algorithm for building abstraction hierarchies automatically. Finally, we empirically demonstrate the advantages of this approach by showing that it works effectively on SSPs of varying size and difficulty.

## Acknowledgements

We gratefully acknowledge the insightful comments by the reviewers. This research was funded in part by the National Science and Engineering Research Council (NSERC), iCore and the Alberta Ingenuity Fund.

## References

Asadi, M., & Huber, M. (2004). State space reduction for hierarchical reinforcement learning. In *FLAIRS*, pp. 509 – 514.

Asadi, M., & Huber, M. (2007). Effective control knowledge transfer through learning skill and representation hierarchies. In *IJCAI*, pp. 2054 – 2059.

Bertsekas, D. P., & Tsitsiklis, J. N. (1996). *Neuro-Dynamic Programming*. Athena Scientific, Belmont, MA.

BioWare Corp. (1998). Baldur's Gate. November 30, 1998.

Blizzard Entertainment (2002). Warcraft III: Reign of Chaos. July 3, 2002.

Bulitko, V., Björnsson, Y., Luštrek, M., Schaeffer, J., & Sigmundarson, S. (2007). Dynamic Control in Path-Planning with Real-Time Heuristic Search. In *ICAPS*, pp. 49–56.

Bulitko, V., Luštrek, M., Schaeffer, J., Björnsson, Y., & Sigmundarson, S. (2008). Dynamic Control in Real-Time Heuristic Search. *JAIR*. In press.

Bulitko, V., Sturtevant, N., Lu, J., & Yau, T. (2007). Graph Abstraction in Real-time Heuristic Search. *JAIR*, *30*, 51–100.

Dean, T., Givan, R., & Leach, S. (1997). Model reduction techniques for computing approximately optimal solutions for Markov decision processes. In *UAI*, pp. 124–131.

Hauskrecht, M., Meuleau, N., Kaelbling, L. P., Dean, T., & Boutilier, C. (1998). Hierarchical solution of Markov decision processes using macro-actions. In *UAI*, pp. 220 – 229.

Isaza, A., Szepesvári, C., Bulitko, V., & Greiner, R. (2008). Speeding up planning in Markov decision processes via automatically constructed abstraction. Tech. rep., Computing Science, U. Alberta. http://www.cs.ualberta.ca/~szepesva/RESEARCH/PRMDP.

Kim, K., & Dean, T. (2003). Solving factored MDPs using non-homogeneous partitions. *Artificial Intelligence*, *147*, 225–251.

McMahan, H. B., & Gordon, G. J. (2005). Fast exact planning in Markov decision processes. In *ICAPS*, pp. 151–160.

Sturtevant, N. (2007). Memory-efficient abstractions for pathfinding. In *AIIDE*, pp. 31–36.

Sutton, R. S., Precup, D., & Singh, S. (1999). Between MDPs and semi-MDPs: a framework for temporal abstraction in reinforcement learning. *Artificial Intelligence*, *112*(1–2), 181–211.